\documentclass[10pt,twocolumn,letterpaper]{article}

\usepackage{iccv}
\usepackage{times}
\usepackage{epsfig}
\usepackage{graphicx}
\usepackage{amsmath}
\usepackage{amssymb}
\usepackage{booktabs}
\usepackage{multirow}
\usepackage{dcolumn}
\usepackage{makecell}

\usepackage[pagebackref=true,breaklinks=true,letterpaper=true,colorlinks,bookmarks=false]{hyperref}

\iccvfinalcopy % *** Uncomment this line for the final submission

\def\iccvPaperID{0000} % *** Enter the ICCV Paper ID here

\ificcvfinal\fi

\begin{document}

%%%%%%%%% TITLE
\title{Efficient DETR: Improving End-to-End Object Detector with Dense Prior}

\author{Zhuyu Yao
\quad
Jiangbo Ai\thanks{This work was done at Megvii Technology.}
\quad
Boxun Li
\quad
Chi Zhang\\
Megvii Technology\\
{\tt\small \{yaozhuyu, liboxun, zhangchi\}@megvii.com
\quad
jiangboml@gmail.com}
}

\maketitle

% Remove page # from the first page of camera-ready.
\ificcvfinal\thispagestyle{empty}\fi

%%%%%%%%% ABSTRACT
\begin{abstract}
The recently proposed end-to-end transformer detectors, such as DETR and Deformable DETR, have a cascade structure of stacking 6 decoder layers to update object queries iteratively, without which their performance degrades seriously.  In this paper, we investigate that the random initialization of object containers, which include object queries and reference points, is mainly responsible for the requirement of multiple iterations. Based on our findings, we propose Efficient DETR,  a simple and efficient pipeline for end-to-end object detection. By taking advantage of both dense detection and sparse set detection, Efficient DETR leverages dense prior to initialize the object containers and brings the gap of the 1-decoder structure and 6-decoder structure.
Experiments conducted on MS COCO show that our method, with only 3 encoder layers and 1 decoder layer, achieves competitive performance with state-of-the-art object detection methods. Efficient DETR is also robust in crowded scenes. It outperforms modern detectors on CrowdHuman dataset by a large margin.
\end{abstract}

%%%%%%%%% BODY TEXT
\section{Introduction}
Object detection is a classic task in computer vision that studies to locate bounding boxes and predicts categories of objects in images. Modern detectors~\cite{FastRCNN, FasterRCNN, YOLO, FocalLoss} generate dense anchors with the sliding window method to establish connections between network predictions and the ground truth. Post-processing methods like non-maximum suppression~(NMS) is used to remove the redundancy of predictions. These hand-crafted components can not achieve the end-to-end object detection. 

Recently, DETR~\cite{DETR} proposed to build an end-to-end framework based on an encoder-decoder transformer~\cite{vaswani2017attention} architecture and bipartite matching, which directly predicts a set of bounding boxes without post-processing. However, DETR requires 10 to 20 times of training epochs than modern mainstream detectors~\cite{FasterRCNN, YOLO, FocalLoss} to converge, and shows relatively low performance in detecting small objects. 
Deformable DETR~\cite{deformabledetr} solves the two problems in the following two aspects: (i). replacing global scoped attention with local spatial attention, which accelerates the convergence of training. (ii). replacing single-scale feature maps with multi-scale feature maps, which significantly improves the performance in detecting small objects.

\begin{figure}[t]
\centering
\includegraphics[width=0.42\textwidth]{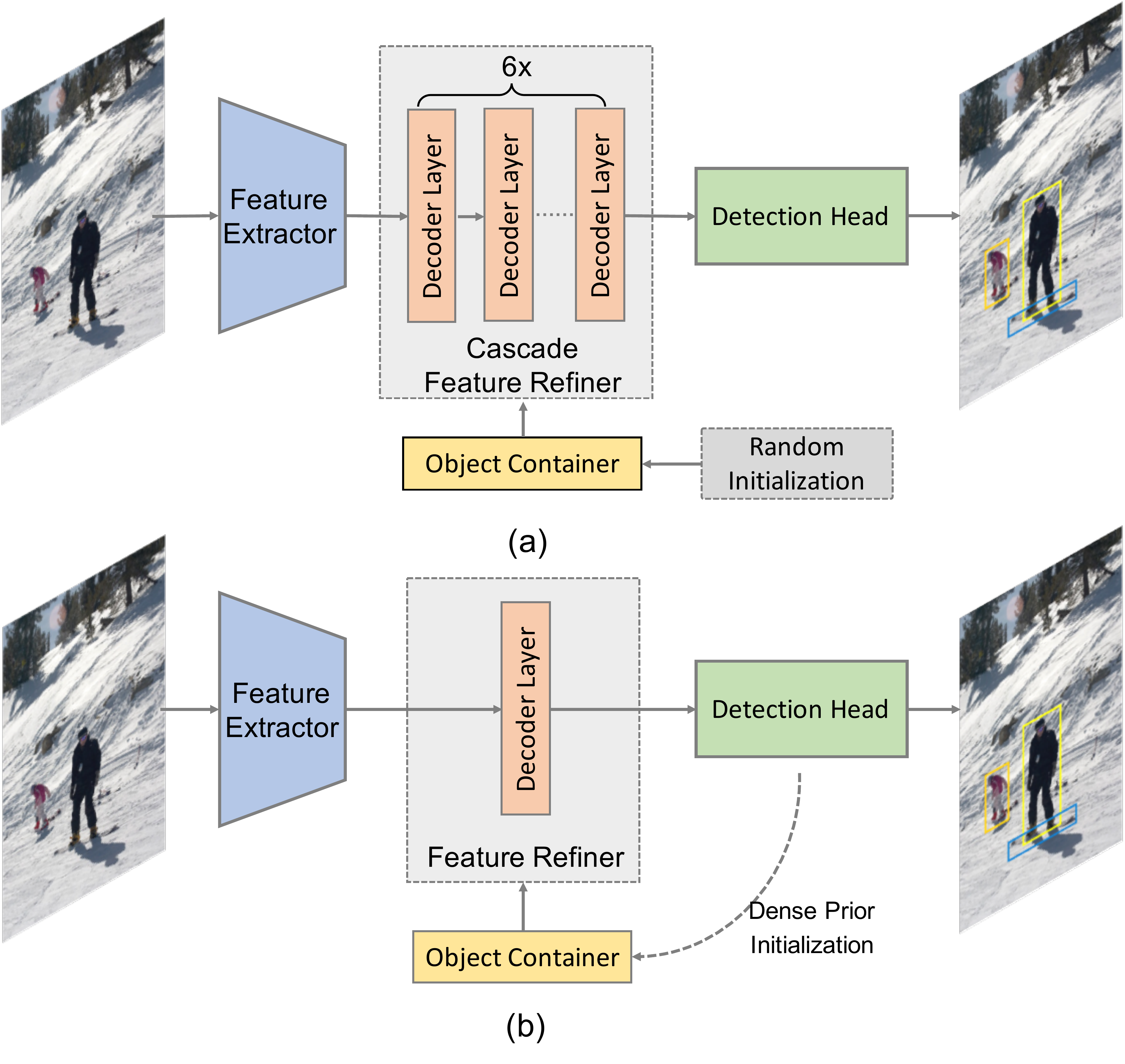}
\caption{Previous end-to-end detectors (a) and ours (b)}
\label{fig:first_page_img}
\vspace{-4mm}
\end{figure}

% object container
The DETR's pipeline can be abstracted as shown in Fig.\ref{fig:first_page_img}(a). We define the \emph{object container} as a container of structured information, which includes different kinds of object features. Object queries and reference points both belong to the object container, since the object queries and the reference points could represent the abstract features and positional information of objects. A set of randomly initialized object containers are fed to a feature refiner to interact with the features extracted from the image. Specifically, 6 decoder layers with cross-attention modules play the role of a cascade feature refiner which updates object containers iteratively. The refined object containers contribute to the final predictions of DETR. Moreover, the feature from the image is extracted from a feature extractor, which has a CNN backbone and 6 encoder layers in DETR. In a word, image and randomly initialized object containers pass through the feature extractor and cascade feature refiner to get final results.
In this pipeline, both DETR and Deformable DETR have a 6-encoder and 6-decoder transformer architecture. We hypothesize that this structure is the key for the DETR series to achieve high accuracy in object detection. 

In this paper, we investigate the components of DETRs for understanding their mechanism. 
We conduct extensive experiments and find that the decoder layer with an extra auxiliary loss contributes most to the performance. 
The transformer decoders iteratively interact object containers with the feature maps. We explore that \textit{the random initialization of object containers in DETRs is mainly responsible for the requirement of multiple refinement times (i.e., decoder layers) and leads slow convergence}.

However, it's hard to directly analyze object queries~\cite{DETR} as they are just a set of abstract features. Thanks to the Deformable DETR, it proposes the reference points~\cite{deformabledetr} for object queries. Reference points are 2-d tensors that represent the guess of box centers.
By visualizing reference points of a trained model, we find that they proves to serve just as anchor points~\cite{FCOS, CenterNet, CornerNet, CornerNet-CenterNet} in anchor-based methods~\cite{FasterRCNN, SSD}.  Furthermore, we report that different initialization of reference points leads to huge performance differences in the 1-decoder structure.
Here comes the question: which initialization is better for object containers in end-to-end detectors? 

In this paper, we propose Efficient DETR, a simple but efficient end-to-end detector~\ref{fig:first_page_img}(b). % based on Deformable DETR. 
Efficient DETR has two parts: dense and sparse. Both parts share the same detection head. In the dense part, a class specific dense prediction based on sliding-windows is performed to generate proposals. Top-K (K=100 in this paper) scored 4-d proposals and their 256-d encoder features are taken as the reference points and object queries. The initialization of object containers is so reasonable that our method is able to achieve better performance and fast convergence with only 1 decoder layer. Specifically, a 3-encoder and 1-decoder structure based on Deformable DETR, could achieve competitive performance (44.2AP in COCO~\cite{COCO}) with a ResNet50~\cite{ResNet} backbone and faster (36 epochs) training. 
%-------------------------------------------------------------------------
\section{Related Work}
\subsection{One-stage and Two-stage Detectors}
Mainstream detectors are mostly based on anchor boxes or anchor points. Anchors are generated at the center of each sliding-window position, which offers candidates for objects. In one-stage detectors, such as YOLO~\cite{YOLO}, SSD~\cite{SSD}, RetinaNet~\cite{FocalLoss}, and FCOS~\cite{FCOS}, the network directly predicts categories and offsets of anchors for the whole feature maps. A one-to-many label assignment rule is usually adopted to compute targets for each anchor. Post-processing such as NMS is used to remove duplicate predictions of objects.
In contrast, two-stage detectors, such as Faster RCNN~\cite{FasterRCNN} and Mask RCNN~\cite{MaskRCNN}, don't directly output final predictions. Instead, they perform class agnostic dense predictions firstly to generate foreground proposals by a Region Proposal Networks~(RPN)~\cite{FasterRCNN, CascadeRCNN, CascadeRPN}. % With NMS and top-k selecting post-processing, a small fixed size of region proposals are picked from the output. 
An ROIPool~\cite{FasterRCNN} or ROIAlign~\cite{MaskRCNN} layer is used to extract features of region proposals from the backbone features. NMS is also used for the final results. The ROIPool and ROIAlign layers solve the feature misalignment problem and refine proposals. As a result, two-stage detectors usually demonstrate better performance than one-stage detectors. 

\subsection{End-to-end Detectors}
End-to-end detectors, such as \cite{wang2020end}, DETR~\cite{DETR}, Deformable DETR~\cite{deformabledetr}, and Sparse RCNN~\cite{sun2020sparse}, don't require extra post-processing stages and perform object detection in an end-to-end framework. DETR is built in an encoder-decoder transformer architecture. Backbone features are passed through 6 encoder layers to extract context information for each position of the feature maps. Object queries, defined as learned positional encodings~\cite{vaswani2017attention, DETR}, are sent to 6 decoder layers to iteratively interact with the encoder features. Object queries have a small size, for example, 100 in DETR. Final results are predicted from object queries by a detection head. A set prediction loss, i.e., the bipartite matching loss, is added to each decoder layer for refinement. Without any hand-crafted components, DETR directly predicts a set of boxes with scores as the final predictions. However, DETR needs a long training process ($\sim$500 epochs) and is not good at detecting small objects. 

To solve these problems, Deformable DETR and Sparse RCNN both propose an effective good solution. In Deformable DETR, multi-scale features are used to help detect small objects. It generates a reference point for each object query, and proposes a deformable attention module~\cite{dai2017deformable, zhu2019deformable, deformabledetr}, to make each reference point only focuses on a small fixed set of sampling points. This modification turns the global connections in the transformer to local connections and significantly raises the converging speed from 500 epochs of DETR to 50 epochs. Deformable DETR is also in a 6-encoder and 6-decoder architecture. 
Sparse RCNN proposed a purely sparse framework for object detection with learnable proposals. A 256-d proposal feature is generated for each learnable region proposal in pair. Similar to the object queries in DETR, proposal features are fed to 6 dynamic instances interactive head~\cite{sun2020sparse} to interact with the region feature from feature pyramid networks~(FPN)~\cite{FPN}. The RoI layer limits the proposal feature to only interact with a few local features and makes Sparse RCNN have a fast converging speed (36 epochs). 
Besides, \cite{sun2020rethinking}, \cite{dai2020up} and \cite{gao2021fast} also studied the convergence problem of DETR in other ways.
%------------------------------------------------------------------------

\begin{table}[t]	
	\centering
    \setlength{\tabcolsep}{1.5mm}
\begin{tabular}{c c c c c c}
\toprule
Encoder & Decoder & Decoding loss &AP &AP$_{50}$ &AP$_{75}$ \\
\midrule
3 &3 &\checkmark &41.5 &60.4 &44.3 \\
3 &1 &\checkmark &32.2 &51.1 &33.7 \\
1 &3 &\checkmark  &39.8 &58.2 &42.5 \\ 
1 &3 & &28.3 &47.4 &29.3 \\
\bottomrule
\end{tabular}
        \vspace{2mm}
	\caption{Encoder vs. Decoder. Experiments are conducted on a Res50 Deformable DETR with 100 proposals and 3$\times$ training schedule.}
    \label{table:compare_enc_dec}
\end{table}

\begin{table}[t]	
	\centering
    \setlength{\tabcolsep}{1.6mm}
\begin{tabular}{l c  c c c c c}
\toprule
Decoder layer  &1 &2 &3 &4 &5 &6 \\
\midrule
AP  &32.1 &39.4 &41.7 &42.4 &42.7 &42.4 \\
AP$_{50}$ &50.8 &58.4 &60.6 &61.5 &61.8 &61.8 \\
AP$_{75}$ &34.0 &41.9 &44.4 &45.5 &45.9 &45.6 \\
\bottomrule
\end{tabular}

        \vspace{2mm}
	\caption{Effect of number of decoder layers in DETR.}
    \label{table:decoder}
    \vspace{-4mm}
\end{table}

\section{Exploring DETR}
\subsection{Revisit DETR}
\noindent{\bf Encoder and decoder.} The DETR series are in a encoder-decoder transformer~\cite{vaswani2017attention} architecture. Both encoder and decoder cascade 6 identical layers. An encoder layer is formed of a multi-head self-attention and a feed-forward network~(FFN), while a decoder layer has an extra multi-head cross-attention layer. The encoder layers play a similar role as convolutions and extract context features from a CNN backbone with multi-head self-attention. In decoders, a set of 256-d object queries interact with encoder features of the whole image and aggregate the information through multi-head cross attention. Auxiliary bipartite matching loss is applied to each decoder layer. Table~\ref{table:compare_enc_dec} illustrates that DETR is more sensitive to the number of decoder layers, which implies decoder is more important than encoder for DETR. Especially, a DETR with 3-encoders and 3-decoders is adopted as our baseline. AP could be decreased by about 9.3 if removing two layers in the decoder. In contrast, removing two layers in the encoder only caused a 1.7 AP drop.

\noindent{\bf Why is the decoder more important than encoder?} Both of them are in a cascade architecture, except that decoder has an extra auxiliary loss for each identical layer. In Table~\ref{table:compare_enc_dec}, we find that the auxiliary decoding loss is the main reason why DETR is more sensitive to the number of decoder layers. The behaviors of the encoder and decoder are tending to the same without the auxiliary loss. We point out that the auxiliary decoding loss introduces strong supervision in updating the query feature, which makes the decoder more efficient. The cascade structure of the decoder refines features with layer-wise auxiliary loss. The more times of iterations, the more efficient auxiliary decoding supervision. 

To further explore the cascade structure of the decoder, we try different numbers of decoder layers. Table~\ref{table:decoder} shows that performance degrades significantly as the times of cascade decrease. There is a huge drop of 10.3 AP between a 6-layer decoder and a 1-layer decoder. It is worth noting that only object queries are updated after each iteration. Object queries are strongly correlated to performance as final predictions are predicted from them by a detection head. However, object queries are randomly initialized at the start of training. We hypothesize that this random initialization does not offer a good initial state, which might be the reason why DETR needs a cascade structure of 6 iterations to achieve competitive performance.

\begin{figure}[t]
\centering
\includegraphics[width=0.4\textwidth]{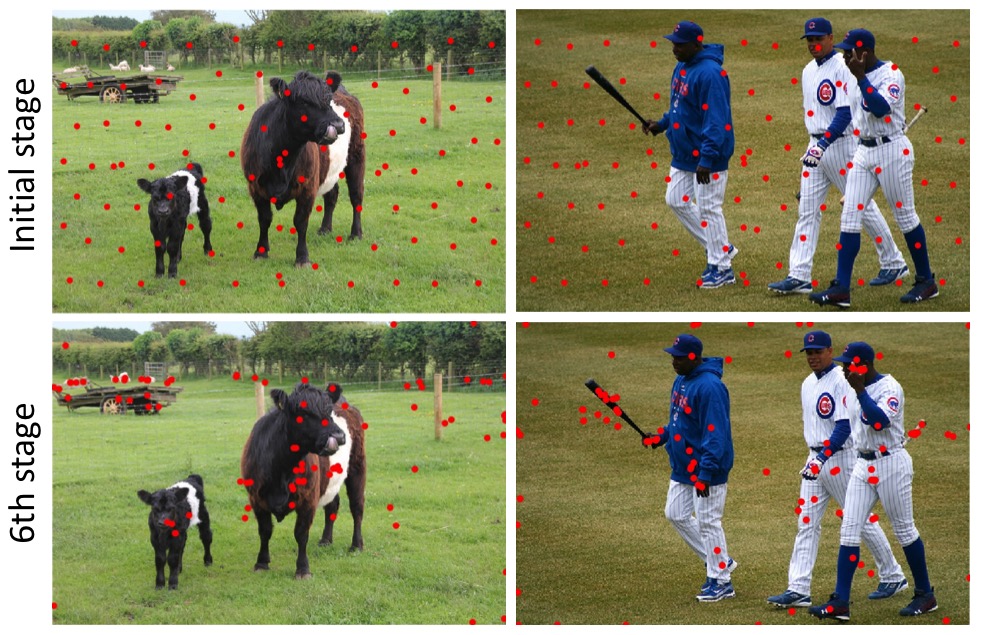}
\vspace{2mm}
\caption{Visualizing the reference points of  Deformable DETR. The reference points at the initial stage are in a distribution similar to the grid of image. After 6 iterations, reference points gather to the centers of foreground.}
\label{fig:rp_learned}
\vspace{-4mm}
\end{figure}

\subsection{Impact of initialization of object containers}
Based on the analysis in the previous section, it is worth studying the initialization of object queries. Object query belongs to the feature information of an object container. The object query is defined as the learned positional embedding, which is a 256-d abstract tensor and hard to analyze. However, we observe that each object query in DETR learns to specialize in certain areas and box sizes with several operating modes. We hypothesize that studying the spatial projection of object query may help understand in an intuitive way. 

Thanks to Deformable DETR~\cite{deformabledetr}, it brings a new component, called reference point that is related to object queries. Reference points are 2-d tensors that represent predictions of box centers and belong to the location information of an object container. In addition, the reference points are predicted from the 256-d object query via a linear projection. 
They could serve as the projection of object query in the 2D space and offer an intuitive presentation of the location information in object query. 
The reference point and object query are updated during iterations of decoders and contribute to the final results. 

\begin{figure}[t]
\centering
\includegraphics[width=0.4\textwidth]{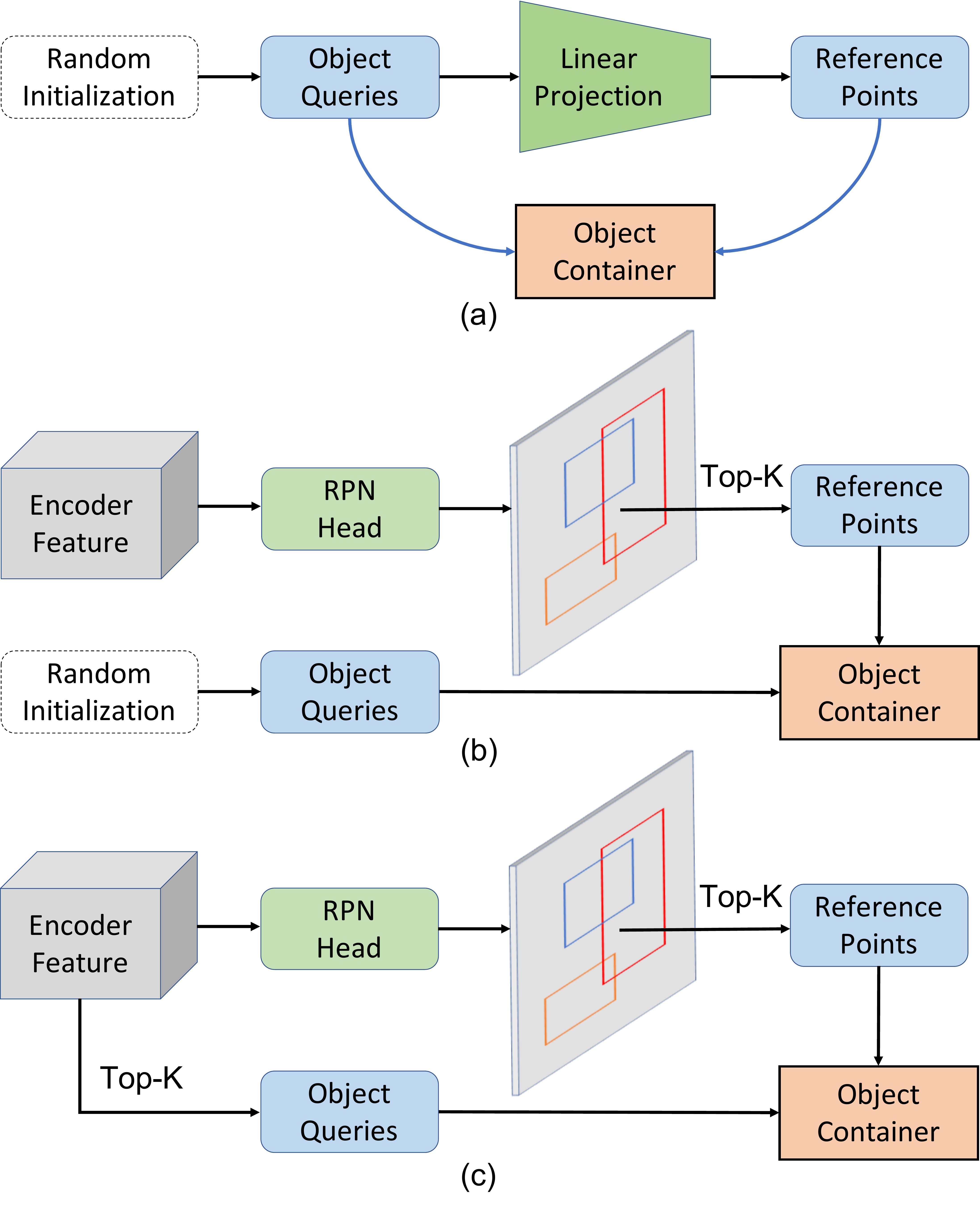}
\vspace{2mm}
\caption{(a). In the original DETR, object queries are randomly initialized. (b). Initializing reference point with top-k scored region proposals from RPN. (c). Based on (b), further initialize object queries with the proposal features of top-k region proposals.}
\label{fig:pipeline3}
\vspace{-4mm}
\end{figure}

Considering that reference points intuitively represent the location information in object queries, we start our exploration from them.
Before passing to decoder layers, reference points are generated via a linear projection over randomly initialized object queries, as shown in Fig.~\ref{fig:pipeline3}(a). We call this process the initialization of reference points. Fig.~\ref{fig:rp_learned} demonstrated the learned reference points of a converged model. The reference points at the initial stage are evenly distributed on the image, covering the whole image area. 
We point out that this initialization is similar to the generation of anchor points~\cite{DenseBox, FCOS, CenterNet, CornerNet, CornerNet-CenterNet} in anchor-based detectors. As the iterative stage increases, reference points gradually gather to the centers of the foreground and finally cover almost all foreground at the last stage. Intuitively, reference points serve as anchor points that locate the foreground and make attention modules focus on a small set of critical sampling points around the foreground. 

After investigating the updates of reference points, we then start our exploration on their initialization, which is the way how the reference points are generated. For the remaining part, we call the initialization of reference points and object queries the initialization of object containers. 

\begin{figure}[t]
\centering
\includegraphics[width=0.48\textwidth]{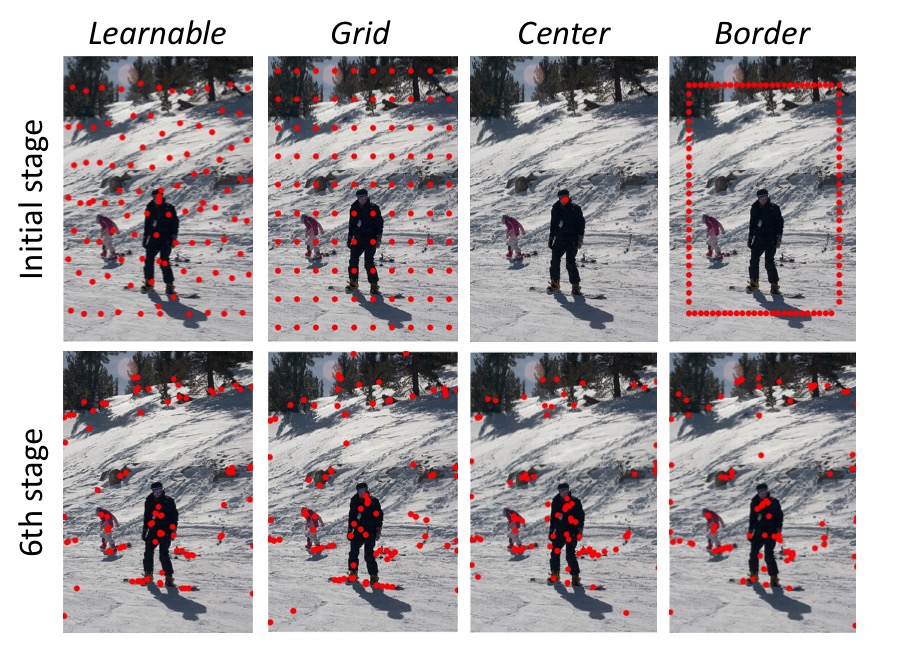}
\caption{Reference points of different initialization. 
Although reference points have different distributions at initial stage, they tends to be in a similar distribution at the final stage.}
\label{fig:rp_compare}
\vspace{-4mm}
\end{figure}

\noindent{\bf Different initialization of reference point.}  In anchor-based detectors, the generation of anchors has a great effect on the model's performance. Mostly, anchors are generated at each sliding-window location, which proves to be a proper initialization for the proposals of objects. The initialization of reference points, which serves as anchor points, may have an effect on the performance of Deformable DETR. We try several different initializations of reference points in both cascade~(6-decoder) and non-cascade~(1-decoder) structure and compare their performance. As shown in Table~\ref{table:dif_rp_init}, different initialization behaves quite differently in non-cascade structures. On the contrary, they lead to a similar performance in a cascade structure. In line with speculation, grid initialization, which generates reference points at the center of the sliding-window, results in approximately equals the performance of learnable initialization. However, the other two kinds of initialization, center, and border, cause a huge drop in accuracy without iterations. For better analysis, we visualize the reference points of different initialization at several stages (Fig.~\ref{fig:rp_compare}). As the stage of iteration increases, their reference points tend to be in the same distribution and locate foreground in a similar pattern at the last stage. 
In a word, different initialization of reference points leads to huge differences in model's performance in non-cascade structure, while cascade structure brings the gap of them by multiple iterations. From another perspective, better initialization of reference points may improve the performance in non-cascade structure.

\noindent{\bf Can we bring the gap of 1-decoder structure and 6-decoder structure with better initialization?} Based on the findings in the previous part, better initialization of reference points might improve the performance, especially for 1-decoder structure. Considering that reference points serve as anchor points, we hypothesize that the anchor prior in mainstream detectors might help us to solve this problem. In modern two-stage detectors, region proposals are generated by RPN in a sliding-window paradigm to offer a set of class agnostic candidates for the foreground. 

RPN is an efficient structure to produce coarse bounding boxes of the foreground with dense prior. As shown in Fig.~\ref{fig:pipeline3}(b), we add an RPN layer on the dense feature from the encoder. The RPN head shares the features of the encoder and predicts the objectness score and offsets of anchors. Top scored bounding boxes are selected as region proposals. We then initialize reference points with the center of region proposals in a non-cascade structure. Table~\ref{table:dif_rp_init} indicates the results, which outperforms other approaches by a large margin, leading to a huge improvement in non-cascade structure. Fig.~\ref{fig:framework} demonstrates the visualization of the method, where reference points at the initial stage get a similar distribution as that of other methods at the last stage. Region proposals initialize reference points in a more reasonable distribution, boosting the accuracy of Deformable DETR without cascade structure.

As we can see in Table~\ref{table:init_rp_ob}, offering the reference point a better initial state with dense prior results in a significant improvement in 1-decoder structure. However, the reference point is just the spatial projection of the object query, while the object query contains extra abstract information of the object container. So, how about initializing the 256-d query feature with dense prior at the same time? 

Intuitively, for each reference point in proposal initialization, we pick its corresponding features from the feature maps, i.e., a 256-d tensor from the encoder, as the initialization of its object query. Our method is illustrated in Fig.~\ref{fig:pipeline3}(c). In Table~\ref{table:init_rp_ob}, our approach further improves 1-decoder structure by 3 AP. Furthermore, only initializing object queries with dense prior and using the original decoder without reference points can also bring a significant improvement to the baseline.

These results indicate that the initial state of the object container, including reference point and object query in Deformable DETR, are highly relevant to the performance of the non-cascade structure. The information of proposals in RPN offering a better initialization with the potential to boost performance by dense priors. Based on our study, we propose Efficient DETR which is able to bring the performance gap between 1-decoder structure and 6-decoder structure.
%-------------------------------------------------------------------------

\begin{table}[t]	
	\centering
    % \centering
% \begin{tabular}{c c c }
% \toprule
% Init. & AP$_{1decoder}$ &AP$_{6decoders}$ \\
% \midrule
% Learnable &32.1 &42.4 \\
% Grid &32.0 &42.7 \\
% Center &21.0 &42.8  \\
% Border &26.0 &42.8\\
% \bottomrule
% \end{tabular}

\setlength{\tabcolsep}{1.5mm}
\begin{tabular}{l c c c c c}
\toprule
Initialization &Learnable &Grid &Center &Border &Dense \\
\midrule
AP$_{\rm 1-dec.}$ &32.1 &32.0 &21.0 &26.0 & 39.0 \\
AP$_{\rm 6-dec.}$ &42.4 &42.7 &42.8 &42.8 & -\\
\bottomrule
\end{tabular}

        \vspace{2mm}
	\caption{Effect of different initialization for reference points. Different initialization results in huge performance gap in 1-decoder structure. However, 6-decoder structure brings the gap of them. 
	}
    \label{table:dif_rp_init}
    \vspace{-4mm}
\end{table}

\begin{table}[t]	
	\centering
    \setlength{\tabcolsep}{1.8mm}
\begin{tabular}{c c c c c c}
\toprule
2-d Ref. &4-d Ref. &Object query &AP & AP$_{50}$ & AP$_{75}$ \\
\midrule
 & & &32.1 & 50.1 & 34.0 \\
\checkmark & & & 39.0 & 57.0 & 42.7 \\
 & &\checkmark &41.1 & 60.8 &45.0 \\
 \checkmark & &\checkmark & 42.0 & 60.5 & 45.6 \\
 & \checkmark &\checkmark &43.0 &60.9 &46.5 \\
\bottomrule
\end{tabular}
        \vspace{2mm}
	\caption{Initializing reference point and object query with dense prior. 2-d Ref. denotes reference points which are 2-d coordinates and initialized with the center points of region proposals. 4-d Ref. denotes reference points which are 4-d bounding boxes and directly initialized with region proposals. Object queries are initialized with 256-d proposal feature vectors from encoder feature maps.}
    \label{table:init_rp_ob}
    \vspace{-4mm}
\end{table}

\section{Efficient DETR}
\begin{figure*}[t]
\centering
\includegraphics[width=0.9\textwidth]{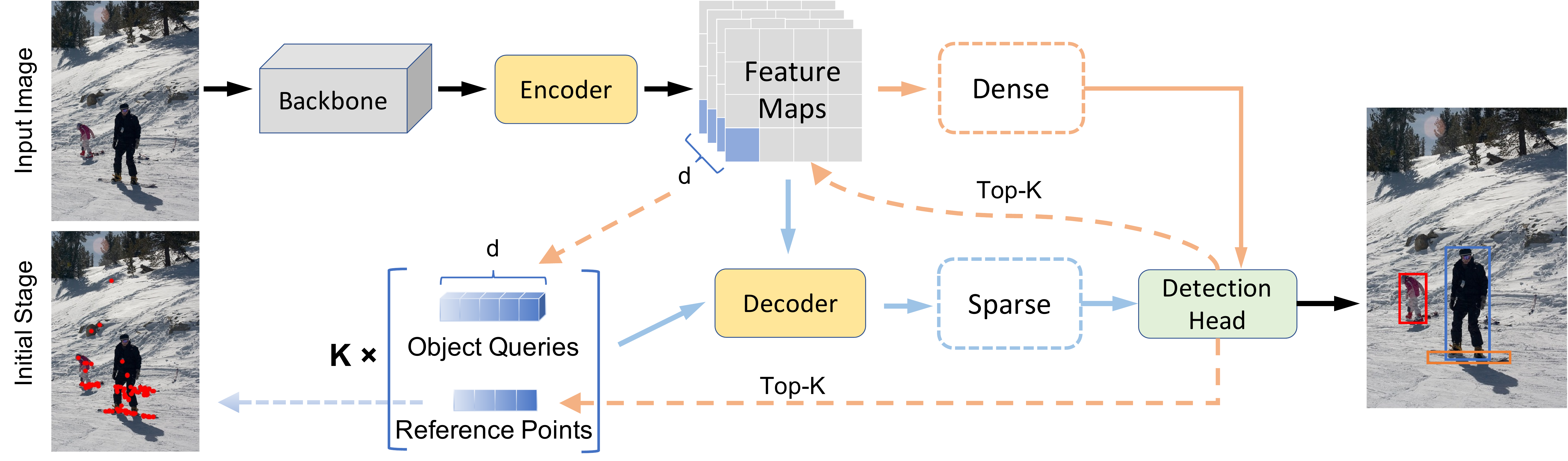}
\vspace{2mm}
\caption{Architecture of Efficient DETR. Top-K denotes the topk selection method. For example, the top scored indices of anchors from the dense part are used to select object queries from encoder, and select reference points from region proposals.}
\label{fig:framework}
\vspace{-4mm}
\end{figure*}

Inspired by the findings in the previous section, we present a simple but efficient framework for object detection, called Efficient DETR. It has 3 encoder layers and only 1 decoder layer, without the cascade structure in the decoder. The architecture is showed in Fig.~\ref{fig:framework}. Efficient DETR is formed of two parts: dense and sparse. 
The dense part does predictions on dense features from the encoder. A top-k selection method is applied to pick a set of proposals from the dense predictions. The 4-d proposals and its 256-d feature from the decoder are taken as the initialization of reference points and object queries. In the sparse part, object containers, the reference points, and object queries initialized with dense prior are fed to a 1-layer decoder to interact with the encoder feature for further refinement. The final results are predicted from the refined object containers. Both parts share the same detection head. All encoder and decoder layers use deformable attention modules proposed in~\cite{deformabledetr}. For the remaining part, we will introduce the details of our network.

\noindent{\bf Backbone.} Following the design of Deformable DETR, we build a backbone with multi-scale feature maps extracted from ResNet. Our backbone has four scales of feature maps with 256 channels. The first three feature maps are extracted from {C3, C4, C5} feature maps of ResNet by a 1$\times$1 stride 1 convolution. While the last feature map is generated via a 3$\times$3 stride 2 convolution on C5. 

\noindent{\bf Dense part.} As mentioned above, the dense part is formed of the backbone, encoder and a detection head. When replacing the encoder with convolutions, it becomes a one-stage detector. Following the two-stage Deformable DETR, anchors are generated for each position at multi-scale feature maps. The base anchor scale is set as 0.05. The detection head predicts C~(C=91 in DETR~\cite{DETR}) category scores and 4 offsets for each anchor. Following~\cite{DETR, deformabledetr, sun2020sparse}, the classification branch is a linear projection layer, and the regression branch is a 3-layer perception with hidden size 256. 

\noindent{\bf Sparse part.} The output of the dense part has a size of encoder feature. We select a set of them by their objectness score, which is defined as the confidence of being a foreground. For each anchor, we take the max category score, predicted by its 256-d encoder feature, as its objectness score. K proposals with the largest scores are picked as reference points. Here, we use reference points as 4-d boxes instead of 2-d centers for more spatial information. As for object queries, they are taken from encoder feature maps. In the dense part, the prediction for each anchor is from a 256-d feature in encoder feature maps. We take this 256-d feature as the proposal feature. The proposal features and reference points appear in pairs and are connected by the same anchor. In our approach, the proposal feature is taken as the object query of its reference point. It makes sense that, firstly, positional encodings are already encoded to the proposal feature when the feature passes through the encoder layers. Secondly, the dense part does the same task as the sparse part, except for the size of their outputs. Intuitively, the feature from the dense part can serve as an initial state of the sparse part. Moreover, given that the tasks of dense part and sparse part are quite similar, these two parts share the same detection head. However, taking each 256-d feature as a training sample, the dense part converges more rapidly than the sparse part because of more training samples per iteration. From this point, dense feature (proposal feature) can serve as a good initialization of sparse feature (object query).

After initialization, object queries are sent to a decoder layer for further enhancement. In traditional detection methods~\cite{DenseBox, FastRCNN, FasterRCNN, YOLO, FocalLoss}, one-stage detectors face feature misalignment, while the two-stage detectors solve this problem by ROIAlign~\cite{MaskRCNN, BorderDet} or ROIPool~\cite{FasterRCNN}. In our sparse part, the misalignment is fixed by the decoder, in which the cross-attention modules enable object queries to aggregate features relevant to it. Final predictions are from these enhanced object queries.

Different from previous DETRs, our number of proposals is dynamically tuned during the training process. Given that the network is not able to predict accurate category scores at the beginning of training, a large number of proposals~($~$300) is set from the start. This mainly ensures almost all foregrounds are covered in the sparse set of proposals. However, a small number of proposals will miss a few hard examples. This results in unstable training. As the training of the network, we decrease the number linearly. The number reduces to 100 in the end. This strategy makes the network efficient, which achieves a comparable accuracy as that trained with 300 proposals, with only 100 proposals.

\noindent{\bf Loss.} Both dense part and sparse part of our framework share the same label assignment rule and loss function. To avoid post-processing like NMS, a one-to-one label assignment rule is adopted. Following~\cite{stewart2016end, DETR, deformabledetr}, we match predictions with ground truth by Hungarian algorithm~\cite{kuhn1955hungarian}. The matching cost is defined the same as the loss function, 
$\mathcal{L}$ = $\lambda_{cls}$$\cdot$$\mathcal{L}_{\mathit{cls}}$$+$$\lambda_{L1}$$\cdot$$\mathcal{L}_{\mathit{L1}}$$+$$\lambda_{giou}$$\cdot$$\mathcal{L}_{\mathit{giou}}$. The loss function is the same as that in~\cite{DETR, deformabledetr, sun2020sparse}. $\mathcal{L}_{\mathit{cls}}$ represents focal loss~\cite{FocalLoss} for classification. $\mathcal{L}_{\mathit{L1}}$ and $\mathcal{L}_{\mathit{giou}}$ represent L1 loss and generalized IoU loss~\cite{GIoU} in for localization. $\lambda_{cls}$, $\lambda_{L1}$ and $\lambda_{giou}$ are coefficients of them. We apply one-to-one label assignment rule in the dense part instead of one-to-many assignment since one-to-many assignment relies on a larger number of proposals to reach similar performance. As shown in section~\ref{assign_section}, 1-to-1 assignment in the dense part lets Efficient DETR achieves high accuracy with a small number of proposals~(100). 

\section{Experiments}
\subsection{Experiment Settings}
\noindent{\bf Dataset.} Our experiments are conducted on the challenging MS COCO benchmark. Models are trained on the COCO train2017 split and evaluated with val2017. We report mAP for performance evaluation following previous research. %using the standard metrics for object detection. 

\noindent{\bf Implementation Details.} We use ImageNet~\cite{ImageNet} pre-trained ResNet-50~\cite{ResNet} as the backbone for Efficient DETR. Multi-scale feature maps from C3 to C6 are used. C6 is generated via a 3 x 3 stride 2 convolution on C5. $M=8$ and $K=4$ are set for deformable attentions, as in \cite{deformabledetr, dai2017deformable, zhu2019deformable}. Models are trained for 36 epochs, and the learning rate is decayed at the 24th epoch by a factor of 0.1. Following~\cite{DETR, deformabledetr, sun2020sparse}, $\lambda_{cls}=2$, $\lambda_{L1}=5$, $\lambda_{giou}=2$. We use Adam optimizer~\cite{adam, jia2016dynamic} with base learning rate of 0.0001, $\beta_{1}=0.9$, $\beta_{2}=0.999$ , and weight decay of 0.0001. 

\begin{table*}
    \centering
    \begin{tabular}{c|cccc|cccccc}
    \toprule
     Model&Epochs&GFLOPs&Params (M)&AP&AP$_{50}$&AP$_{75}$&AP$_S$&AP$_M$&AP$_L$  \\
     \midrule
     DETR-R50~\cite{DETR}&500 &86& 41& 42.0&62.4 & 44.2& 20.5& 45.8& \textbf{61.1} \\
     DETR-DC5-R50~\cite{DETR}&500&187& 41&43.3 & 63.1 & 45.9 & 22.5 & 47.3 & 61.1 \\
     Faster RCNN-FPN-R50~\cite{DETR}&36& 180&  42&40.2 & 61.0 & 43.8& 24.2& 43.5& 52.0 \\
     Deformable DETR-R50~\cite{deformabledetr} &50&173 & 40 &43.8 &62.6 &47.7 &26.4 &47.1 &58.0 \\
     TSP-FCOS-R50 ~\cite{sun2020rethinking}& 36 &189& -&43.1 &62.3 &47.0 & 26.6& 46.8 & 55.9\\
     TSP-RCNN-R50~\cite{sun2020rethinking} & 36 &188& -&43.8 &63.3 &48.3 & \textbf{28.6}& 46.9 & 55.7\\
     SMCA-R50~\cite{gao2021fast} &50 & 152& 40 &43.7 &\textbf{63.6}&47.2&24.2 &47.0 &60.4 \\
     Sparse R-CNN-R50~\cite{sun2020sparse} &36 &- &- &44.5 &63.4 &48.2 &26.9 & 47.2 & 59.5 \\
     \midrule
     Efficient DETR-R50 &36 &159 &32 &44.2 &62.2 &48.0 &28.4 &47.5 &56.6 \\
     Efficient DETR$^\ast$-R50 &36 &210 &35 &\textbf{45.1} &63.1 &\textbf{49.1} &28.3 &\textbf{48.4} &59.0 \\
     \midrule
     DETR-R101~\cite{DETR} &500&152 & 60 &43.5&63.8 & 46.4& 21.9& 48.0& 61.8 \\
     DETR-DC5-R101~\cite{DETR} &500&253 & 60 &44.9 & 64.7 & 47.7 & 23.7 & \textbf{49.5} & \textbf{62.3} \\
     Faster RCNN-FPN-R101~\cite{DETR}&36&256& 60 &42.0 &62.1 &45.5 &26.6 &45.4 &53.4 \\
     TSP-FCOS-R101~\cite{sun2020rethinking} & 36 &255& -&44.4 &63.8 &48.2 & 27.7& 48.6 & 57.3\\
     TSP-RCNN-R101~\cite{sun2020rethinking} & 36 &254& -&44.8 &63.8 &49.2 & \textbf{29.0}& 47.9 & 57.1\\
     SMCA-R101~\cite{gao2021fast} & 50 & 218 & 58 & 44.4 & \textbf{65.2} & 48.0 & 24.3 & 48.5 & 61.0 \\
     Sparse R-CNN-R101~\cite{sun2020sparse} &36 & - & - &45.6 &64.6 &\textbf{49.5} &28.3 & 48.3 & 61.6 \\
     \midrule
     Efficient DETR-R101 & 36 &239 &51 &45.2 &63.7 &48.8 &28.8 &49.1 &59.0 \\
     Efficient DETR$^\ast$-R101 & 36 &289 &54  &\textbf{45.7} &64.1 &49.5 &28.2 &49.1 &60.2 \\
     \bottomrule
\end{tabular}
        \vspace{2mm}
    \caption{Comparison with modern object detectors on COCO 2017 validation set. $^\ast$ denotes Efficient DETR is built in a 6-encoder and 1-decoder structure}
    \label{tab:overview}
    \vspace{-4mm}
\end{table*}

\subsection{Main Result}
Our main results are shown in Table~\ref{tab:overview}. All models are evaluated on COCO 2017 validation set. Efficient DETR is compared to the mainstream detector, Faster RCNN, and other state-of-the-art end-to-end detectors such as the DETRs and Sparse RCNN. From the table, we can see that our Efficient DETR based on ResNet50 achieves 44.2 AP with 36 epochs training, outperforming Faster RCNN and most end-to-end detectors with fewer FLOPs and parameters. On one hand, Efficient DETR remains the fast-converging characteristic of Deformable DETR. It has a 10 times faster convergence speed than the original DETR (36 epochs vs 500 epochs). On the other hand, Efficient DETR outperforms Deformable DETR by 0.4 AP with a simpler structure (3-encoder and 1-decoder vs 6-encoder and 6-decoder) and fewer training epochs (36epochs vs 50 epochs). Compared to other end-to-end models, our Efficient DETR is also comparable in performance but more efficient. There is only a 0.3 drop in AP (44.2 AP vs 44.5 AP) compared to the state-of-the-art Sparse RCNN. However, the parameters of Efficient DETR are 20 percent fewer than that of most models (32M vs 40M). And Efficient DETR only needs a small number of 100 proposals to achieve this performance, which also contributes to its efficiency. In contrast,  300 proposals and 700 proposals are required in \cite{deformabledetr} and \cite{sun2020sparse}, respectively. To further improve the Efficient DETR, more encoder layers are stacked to enhance its dense feature. It achieves the-state-of-the-art in 45.2 AP, with a 6-encoder and 1-decoder structure, which still has fewer parameters than other end-to-end models, thanks to the reasonable design of its framework.

\subsection{Ablation Study}
Ablation studies are performed to analyze the components of Efficient DETR. Models in this part are based on Deformable DETR with iterative bounding box refinement, an optimized version that iteratively updates reference points after each decoder layer. If no special instructions, we use ResNet50, a 3-layer encoder, 1-layer decoder, 100 proposals, and a 3$\times$ (36 epochs) training schedule.

\noindent{\bf Detection head for the dense part.} In Efficient DETR, both dense and sparse parts share the same detection head, which predicts C (C$=$91~\cite{DETR, deformabledetr} in DETR) category scores for each anchor. We take the max category score of each anchor as its objectness score, which denotes the confidence of being a foreground. While in mainstream detectors~\cite{FasterRCNN, MaskRCNN, R-FCN}, RPN directly predicts objectness score by a binary class-agnostic classifier. The top-scored proposals and their features are selected as the initialization of object containers.
The selected proposals for initialization of object containers are decided by the top-K max scores of all anchors, which are foreground score for class agnostic and max score for class-specific. An extra detection head is built to explore the effect of class agnostic and class-specific on the initialization of object containers. 
We conduct ablation studies based on the dense part. 
Table~\ref{table: obscore_sharehead} shows class-specific could achieve a better accuracy. 
We hypothesize that compared to class agnostic, class-specific brings more supervise to the dense part. Class-specific enhances the information of categories in 256-d encoder features, thus offering object queries a better initial state in the sparse part. Furthermore, Table~\ref{table: obscore_sharehead} shows sharing the detection head of both parts does not harm performance.

\noindent{\bf Number of proposals.} In \cite{sun2020sparse}, number of proposals largely effects model's performance. In contrast,  Table~\ref{table:num_proposals} shows that increasing proposal numbers from 100 to 1000 results in tiny improvement (0.2 AP) in our work. We hypothesize that the dense part in Efficient DETR, which checks the whole area of the image to find foreground, makes it not sensitive to the number of proposals. Furthermore, our proposed method, decreasing the number of proposals linearly from a large number to 100 as training progresses, proves to bring the gap of DETR with the different number of proposals. Training from a larger number of proposals makes the training process more stable. Since at the beginning of training, the model easily misses proposals for a few objects, which might result in instability of label assignment. As the training goes on, the model is able to give the most foreground a high score. And there is no need to cover them by gathering lots of proposals.

\begin{table}[t]	
	\centering
    \setlength{\tabcolsep}{1.6mm}
\begin{tabular}{c c c c c c}
\toprule
Agnostic &  Specific & Share Head &AP & AP$_{50}$ & AP$_{75}$ \\
\midrule
\checkmark & & &43.0 &60.9 &46.5 \\
\checkmark &\checkmark &  &43.8  &62.0  &47.5 \\
 &\checkmark & &43.8 &61.9 &47.4 \\
 &\checkmark &\checkmark &43.8 &61.9 &47.4  \\
\bottomrule
\end{tabular}
        \vspace{2mm}
	\caption{Class agnostic vs. class specific for the dense head.}
    \label{table: obscore_sharehead}
    \vspace{-1mm}
\end{table}

\begin{table}[t]	
	\centering
    \setlength{\tabcolsep}{2.3mm}
\begin{tabular}{l c  c c  c}
\toprule
Decoder Layer  &1 &2 &3  &6 \\
\midrule
AP & 44.2 & 44.9 & 44.7 & 44.1\\
AP$_{50}$ & 62.5 & 63.4 & 63.7 & 63.7\\
AP$_{75}$ & 47.8 & 48.4 & 48.2 & 47.7\\
\bottomrule
\end{tabular}

    	\vspace{2mm}
	\caption{Stacking decoder layers on Efficient DETR.}
    \label{table:further_improve}
    \vspace{-5mm}
\end{table}

\label{assign_section}
\noindent{\bf Label assignment in the dense part.} Both dense and sparse parts in Efficient DETR use the one-to-one label assignment rule. Given that modern dense detectors~\cite{FCOS, FocalLoss, YOLOv3, ATSS, PAA, zhu2020autoassign, YOLO9000, NoisyAnchors, FreeAnchor, RefineDet}, apply one-to-many label assignment rules, we have a try on the dense part. 1-to-N (N = \{1, 5, 10\}) assignment are evaluated. The proposals, whose IoU with the ground truth are sorted top N (N $>$ 1), are assigned as positive sample. As shown in Table~\ref{table:assign}, one-to-many assignment leads to performance degrade. As the N becomes large, the AP drops significantly (35.6 AP with 1 vs 10 assignment). By increasing the number of proposals when models are evaluated, we find that models trained with one-to-many assignment rules show obvious improvement. For example, models trained with 1 vs 10 assignments have an improvement of 2.2 AP (37.8 vs 35.6). However, models trained with 1 vs 1 assignment show no improvement. We hypothesize that the dense part trained with one-to-many assignment needs more proposals to get a reach high accuracy. Since 1-to-N assignment leads to duplicate predictions of the foreground. In this case, a large number of proposals are needed to cover proposals for all the objects in the image. 

\noindent{\bf Number of encoders and decoders.} Since Efficient DETR has been achieved high accuracy in COCO with a 3-encoder and 1-decoder structure. Table~\ref{tab:overview} shows that stacking another 3 encoder layers improves performance by a margin of 0.9 AP. However, as shown in Table~\ref{table:further_improve}, stacking decoder layers after Efficient DETR brings slight improvement. Instead, the model's performance even degrades after two iterations. It indicates that cascading decoder layers to Efficient DETR is not necessary.

\subsection{Evaluation on CrowdHuman} We evaluate our method on the CrowdHuman dataset~\cite{shao2018crowdhuman} to verify the robustness of Efficient DETR in crowded scenes. As shown in Table~\ref{table: crowdhuman}, Efficient DETR outperforms other detectors by a large margin, \textit{e.g.}, 4 AP and 5 mMR gains over the Deformable DETR. Especially, our approach achieves this with only 100 proposals. More proposals (\textit{e.g.} 400) result in a rise in mMR. We hypothesize that the dense part, which enables Efficient DETR to cover almost all foregrounds in crowded scenes, plays a significant role in its robustness. 1-to-1 assignment in dense part lets our model perform well in crowded scenes with a few proposals. A large number of proposals may bring more false positives to the final results. It shows that our approach has a strong generalization ability and can handle object detection in general scenes efficiently. 

\begin{table}[t]	
	\centering
    % \setlength{\tabcolsep}{1.8mm}
% \begin{tabular}{c c c c c}
% \toprule
% Proposals & Linearly Decrease &AP &AP$_{50}$&AP$_{75}$ \\
% \midrule
% 100 &\checkmark & & &   \\
% 300 & & & &   \\
% 500 & & & &   \\
% 1000 & & & &  \\

% \bottomrule
% \end{tabular}

\setlength{\tabcolsep}{1.8mm}
\begin{tabular}{l c c c c}
\toprule
Proposals &100 &300 &500 &1000 \\
\midrule
Fixed &43.8 &44.2 &44.1 &44.0 \\
Linear Decrease &- &44.2 &44.0 &44.0 \\

\bottomrule
\end{tabular}
    	\vspace{2mm}
	\caption{Effect of the proposal numbers. Fixed denotes model is trained in a fixed number of proposals during the whole training process. Instead, Linear Decrease denotes the number of proposals linearly decreases in the training process. }
    \label{table:num_proposals}
    \vspace{-1mm}
\end{table}

\begin{table}[t]	
	\centering
    \setlength{\tabcolsep}{1.8mm}
\begin{tabular}{l c c c c}
\toprule
Proposals &100 &300 &500 &1000 \\
\midrule
1-to-1 &43.8 &43.9 &43.9 & 43.8\\
1-to-5 &38.0 &39.2 &39.4 &39.5 \\
1-to-10 &35.6 &37.4 &37.6 &37.8 \\

\bottomrule
\end{tabular}
        \vspace{2mm}
	\caption{Impact of different assignment in the dense part.}
    \label{table:assign}
    \vspace{-5mm}
\end{table}

\section{Conclusion}
In this work, we conduct experiments on DETR to explore its components. Explorations on object containers offer us a better understanding of reference points, which serve as the anchor points of object queries. We point out that the random initialization of object containers is the main reason why modern end-to-end detectors need multiple iterations to reach high accuracy. With the dense prior in anchor-based methods, we propose a simple pipeline, Efficient DETR, for end-to-end object detection. Efficient DETR combines the characters of both dense detection and set detection and achieves high performance with a fast convergence speed. We hope our work could inspire designs of more simple and efficient object detectors.

\begin{table}[t]	
	\centering
    \setlength{\tabcolsep}{1.8mm}
\setlength{\tabcolsep}{1.0mm}
\begin{tabular}{l c c c c}
\toprule
Method &Proposals &AP$_{50}$ &mMR &Recall \\
\midrule
Faster-RCNN~\cite{FasterRCNN} &- &85.0 &50.4 &90.24 \\
RetinaNet~\cite{FocalLoss} &- & 81.7 &57.6 & 88.6 \\
FCOS~\cite{FCOS} &- &86.1 &55.2 &\textbf{94.3} \\
ATSS~\cite{ATSS} &- &\textbf{87.1} &\textbf{50.1} &94.0 \\
% AdaptiveNMS~\cite{liu2019adaptive} &- &84.7 &\textbf{49.7} &91.27 \\
\midrule
POTO+3DMF+Aux~\cite{wang2020end} &- &89.2 &49.6 &96.6 \\
DETR~\cite{DETR} &400 &66.12 &80.62 &- \\
Deformable DETR~\cite{deformabledetr} &400 &86.74 &53.98 &92.51 \\
Efficient DETR &400 &90.68 &49.80 &\textbf{97.99} \\
Efficient DETR &100 &\textbf{90.75} &\textbf{48.98} &97.94 \\

\bottomrule
\end{tabular}
        \vspace{2mm}
	\caption{Results on CrowdHuman.}
    \label{table: crowdhuman}
    \vspace{-4mm}
\end{table}

{\small
\bibliographystyle{ieee_fullname}
\bibliography{Efficient_DETR}
}

\end{document}